\documentclass[a4paper]{article}

\usepackage{dx}
\usepackage[T1]{fontenc}
\usepackage{ae,aecompl}
\usepackage{amsfonts}
\usepackage{amsmath}
\usepackage{amsthm}
\usepackage{times}
\usepackage{subfiles}
\usepackage{graphicx}
\usepackage{float}
\usepackage{subcaption}
\usepackage{braket}
\usepackage{xcolor}
\usepackage{mdframed}
\usepackage{url}

\mdfdefinestyle{frameStyle}{%
    roundcorner=5pt,
    backgroundcolor=gray!10!white}

\newcommand{\corr}[0]{\operatorname{corr}}
\newcommand{\cov}[0]{\operatorname{cov}}

\renewcommand{\vec}[1]{\mathbf{#1}}

\title{Data-Driven Robot Fault Detection and Diagnosis Using Generative Models: A Modified SFDD Algorithm}
\author%
{%
    Alex Mitrevski \and
    Paul G. Pl{\"o}ger\\
    Hochschule Bonn-Rhein-Sieg, Sankt Augustin, Germany \\
    e-mail: \{aleksandar.mitrevski, paul.ploeger\}@h-brs.de
}

\begin{document}

\maketitle

    \begin{abstract}
        This paper presents a modification of the data-driven sensor-based fault detection and diagnosis (SFDD) algorithm for online robot monitoring. Our version of the algorithm uses a collection of generative models, in particular restricted Boltzmann machines, each of which represents the distribution of sliding window correlations between a pair of correlated measurements. We use such models in a residual generation scheme, where high residuals generate conflict sets that are then used in a subsequent diagnosis step. As a proof of concept, the framework is evaluated on a mobile logistics robot for the problem of recognising disconnected wheels, such that the evaluation demonstrates the feasibility of the framework (on the faulty data set, the models obtained $88.6\%$ precision and $75.6\%$ recall rates), but also shows that the monitoring results are influenced by the choice of distribution model and the model parameters as a whole.
    \end{abstract}

    \section{Introduction}
    \label{sec:introduction}

    To increase the autonomy and adaptivity of robots, learning-based fault detection and diagnosis (FDD) methods represent a viable alternative to classical model-based algorithms since they minimise the need for accurate, manually specified behavioural models, which are often impractical to obtain. In principle, the reliable application of learning to FDD does require injecting some knowledge about the system into the learning process in order to increase the learning efficiency and improve the model's contextual awareness.

    In \cite{khalastchi2018}, Khalastchi and Kalech describe two versions of a data-driven FDD algorithm, called sensor-based fault detection and diagnosis (SFDD), that uses information about the structural model of a system as well as the correlations between sensor measurements in order to detect and subsequently diagnose robot faults. The general idea behind this method is to find the pairs of correlated sensors in a system and then, by monitoring manually specified \emph{data modes}\footnote{In \cite{khalastchi2018}, these are referred to as \emph{patterns}.}(such as stuck at or drifting values), look for violations of those correlations, which are taken to be indications of a fault. Faults detected in this manner can then be diagnosed in a subsequent step.

\begin{figure}[t]
        \centering
        \includegraphics[scale=0.23]{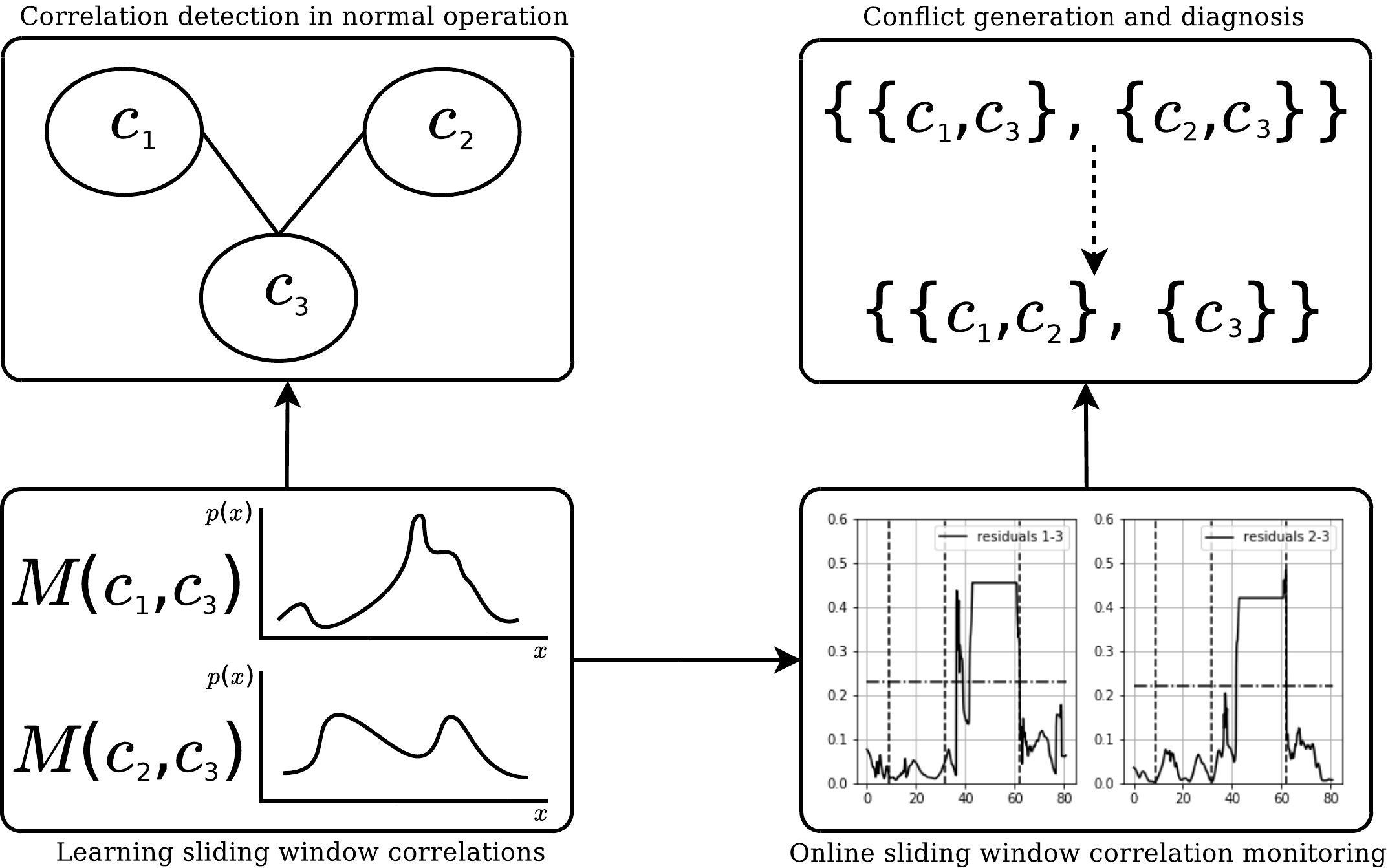}
        \caption{Overview of our learning-based FDD schema}
        \label{fig:schema_overview}
    \end{figure}

    The practical usefulness of the SFDD algorithm is significantly affected by the choice of data modes that are to be monitored during the operation of a robot, since an incomplete or suboptimal choice of modes leads to either undetected correlation violations or a large number of false positive detections.\footnote{Our implementation of the SFDD method, along with real-robot data from a KUKA youBot that demonstrate the problem, can be found at \url{https://github.com/alex-mitrevski/SFDD}} To address this issue, we present a modification of the mode monitoring method in \cite{khalastchi2018}, such that we replace the manually specified modes with \emph{models of pairwise sliding window correlations}, namely we learn a probability distribution of sliding window correlations between the measurements of the correlated sensor pairs. Each such distribution is represented by a generative model, which is used in a residual generation scheme during online operation; these residuals are then used for conflict set generation and diagnosis. Our methodology is summarised in Fig. \ref{fig:schema_overview}.

    We use Restricted Boltzmann Machines (RBMs) for distribution modelling due to their flexibility and extensibility \cite{fischer2012,zhang2019}. For each model, a residual is generated by sampling from the distribution given a window of measurement correlations and comparing the sample with the observation. A residual higher than a model-specific threshold indicates a fault in one of the components and is used to create a conflict set. For finding diagnoses given the conflict sets, we use the HS-DAG algorithm \cite{greiner1989} as implemented in \cite{quaritsch2014}. The feasibility of our framework is analysed for a mobile robot that is used in a logistics application \cite{mitrevski2018}. We additionally compare RBMs to Gaussian Mixture Models (GMMs) on the same data to illustrate the conceptual independence on the choice of generative model. As shown in section \ref{sec:experiments}, our representation is not constrained to RBMs, as other generative models can be used as well, although the choice of model does have an influence on the quality of the generated residuals.

    We organise this paper as follows. Section \ref{sec:related_work} discusses related work in learning-based FDD; the SFDD algorithm is described in more detail in section \ref{sec:preliminaries}, along with some preliminaries on restricted Boltzmann machines; section \ref{sec:learning} presents our modification to the SFDD method, while section \ref{sec:experiments} shows results of preliminary experiments performed with a mobile robot platform; finally, section \ref{sec:discussion} concludes the paper.

    \section{Related Work}
    \label{sec:related_work}

    Generative models, in particular restricted Boltzmann machines, have been used for anomaly detection in various contexts. Wulsin et al. \cite{wulsin2010} use a deep belief network (which is made up of restricted Boltzmann machines) for finding anomalies in electroencephalography (EEG) waveforms; deep belief networks are also shown to be more reliable than one-class support vector machines in this context. Chopra and Yadav \cite{chopra2018} apply restricted Boltzmann machines for feature extraction from acoustic signals; these features are then used for detecting faults in a combustion engine. A similar application of restricted Boltzmann machines to the problem of bearing fault detection is discussed in \cite{chen2016} and \cite{zhang2019}.

    More recent generative models, such as Generative Adversarial Networks (GANs) and Variational Autoencoders (VAEs), have also been applied to anomaly detection. In \cite{dimattia2019}, Di Mattia et al. compare various GAN models for anomaly detection, mostly on the task of detecting visual anomalies. Zhang and Chen \cite{zhang2019_2} on the other hand present a model that combines a VAE and an LSTM (long short-term memory) network, which is used for  detecting anomalies in electrocardiogram (ECG) data. Since our aim is to learn pairwise correlation models, RBMs are advantageous since they are easier to train than VAEs and GANs, and, as shown in section \ref{sec:experiments}, can work reliably even when a smaller model is used.

    Learning-based fault detection and diagnosis have also been successfully applied in robotics for addressing different aspects of the problem. Christensen et al. \cite{christensen2008} use a time-delay neural network for fault detection, where fault injection is used for training data collection; similar to this work, we learn data models from sliding windows, but we learn correlations instead of direct observations. In \cite{golombek2010}, Golombek et al. learn the distribution of time intervals between occurrences of event pairs; these distributions are then used for assigning scores to sequences of observations, which allows detecting anomalies in the observations. In a similar manner, Li and Parker \cite{li2009} learn a state transition diagram on features extracted by clustering sensor observations, such that unlikely transitions or unusually long periods of time in a single state are used as indications of anomalies. Fox et al. \cite{fox2007} use learned hidden Markov models for execution failure detection, where the states of the model are obtained by clustering observations in particular executions of a given task. We consider these models to be complementary to ours since they analyse executions rather than the health of components as such, which is what we focus on in this paper.

    \section{Preliminaries}
    \label{sec:preliminaries}

    \subsection{Sensor-Based Fault Detection and Diagnosis (SFDD)}
    \label{sec:preliminaries_sfdd}

    The SFDD method presented in \cite{khalastchi2018} comes in two variants, the so-called basic and extended versions. In this paper, we focus on the extended approach, where we are given a structural model of a system, a set of sensors $S$, and a set $P$ of data modes that measurements might follow, such as drift or zero slope. The algorithm operates in two phases - an offline and an online phase. In the offline phase, we are given a matrix $X$ of size $m \times n$, where $m$ is the number of sensors and $n$ is the length of a sequence of consecutive sensor measurements collected during fault-free system operation. The objective is to find a set $C$ of \emph{correlated sensors}
    \begin{equation}
        C = \{ (S_i, S_j) \; | \; 1 \leq i, j \leq m, i \neq j, \corr(S_i, S_j) = 1 \}
        \label{eq:correlated_sensors}
    \end{equation}
    where
    \begin{equation*}
        \corr(S_i, S_j) =
        \left\{
            \begin{array}{ll}
                1 & \text{if } S_i \text{ and } S_j \text{ are correlated} \\
                0 & \text{otherwise}
            \end{array}
        \right.
    \end{equation*}
    along with pairs of modes that are observed together in windows of $X$; the mode identification step thus takes into consideration scenarios in which the correlation between sensors is temporarily lost and the sensors follow different modes. In the online phase, the modes exhibited by each sensor $S_i$ are continuously monitored and compared against those of the correlated sensors; if a mode exhibited by $S_i$ is not observed for any of the correlated sensors that belong to independent components according to the structural graph, $S_i$ is considered faulty.

    \subsection{Restricted Boltzmann Machines}
    \label{sec:preliminaries_rbms}

    As mentioned before, we use Restricted Boltzmann Machines (RBMs) as generative data models for anomaly detection. While a detailed treatment of RBMs is not in the scope of this paper, we present a brief summary of RBMs and their working mechanism here, drawing much of the content and notation from \cite{koller2009,fischer2012,hinton2012}.

	An RBM is an undirected graphical model given in the form of a fully-connected bipartite graph. One of the graph's layers, called the \emph{visible layer}, has a number of units equal to the dimensionality of the model's input data, which is denoted $|V|$, and the other layer is called the \emph{hidden layer} and has dimensionality that is denoted $|H|$. Being an undirected model, and RBM encodes a Gibbs distribution of the form \cite{koller2009}
    \begin{equation*}
        P(v,h) = \frac{1}{Z}e^{-E(v,h)}
    \end{equation*}
    where $E(v,h)$ is an \emph{energy function} that is given as
    \begin{equation*}
        E(v,h) = -\sum_{i=1}^{|H|}\sum_{j=1}^{|V|}w_{ij}h_iv_j - \sum_{j=1}^{|V|}a_jv_j - \sum_{i=1}^{|H|}b_ih_i
    \end{equation*}
    and $Z$ is called a \emph{partition function} and acts as a normalising constant, which is calculated as
    \begin{equation*}
        Z = \sum_{j=1}^{|V|}\sum_{i=1}^{|H|}e^{-E(v_j,h_i)}
    \end{equation*}
    In the above equations, $w_{ij}$ is a connection weight between the $i$-th hidden unit and the $j$-th visible unit, $a_j$ is a bias term of the $j$-th visible unit, and $b_i$ is a bias term of the $i$-th hidden unit.

    Assuming binary units in the network, the conditional probability of a unit being equal to one can be written as
    \begin{equation*}
        \begin{aligned}
            p(h_i=1|v) = \sigma\left(\sum_{j=1}^{|V|}w_{ij}v_j + b_i\right) \\
            p(v_j=1|h) = \sigma\left(\sum_{i=1}^{|H|}w_{ij}h_i + a_j\right)
        \end{aligned}
    \end{equation*}
    where $\sigma$ is the logistic function
    \begin{equation*}
        \sigma(x) = \frac{1}{1 + e^{-x}}
    \end{equation*}

    Training an RBM consists of learning the distribution represented by the training samples, which entails maximising the likelihood, or the log-likelihood in practice, of the model parameters given the training data. It can be found that in order to maximise the (log-)likelihood, summations over the visible variables are required; however, these will generally be difficult or impossible to compute. As a result of that, these terms are approximated by sampling from the distribution encoded by the RBM and running a Markov chain for a few iterations. Such an approximation can be performed by an algorithm called \emph{Contrastive Divergence} (CD-$k$), where $k$ is the number of iterations of the sampling algorithm. Once a sample has been obtained, the updates of the weights and biases can be performed using the following update terms:
    \begin{equation}
        \begin{array}{ll}
            \Delta w_{ij} = & p(h_i=1|v^0)v_j^0 - p(h_i=1|v)v_j \\
            \phantom{\Delta w_{ij}} = & \Braket{v_ih_j}_{data} - \Braket{v_ih_j}_{model}
        \end{array}
    \end{equation}

    \begin{equation}
        \Delta a_j = v_j^0 - v_j^k \Braket{v_j}_{data} - \Braket{v_j}_{model}
    \end{equation}

    \begin{equation}
        \begin{array}{ll}
            \Delta b_i = & p(h_i = 1|v^0) - p(h_i = 1|v) \\
            \phantom{\Delta b_i} = & \Braket{h_j}_{data} - \Braket{h_j}_{model}
        \end{array}
    \end{equation}
    Here, $v$ is a sample obtained using CD, $v^0$ is the current training sample, and $\Braket{}$ denotes an average log-likelihood: $\Braket{}_{data}$ is the likelihood obtained by assigning training data to the visible layer, while $\Braket{}_{model}$ is the likelihood resulting from the model samples.

    As discussed in \cite{hinton2012}, CD-$1$ can provide sufficiently good estimates of the likelihood in practice. For that reason, we use this version of the algorithm rather than a more expensive one that would approximate the likelihood more closely.

    \section{Learning Generative Dependency Models}
    \label{sec:learning}

    In this section, we first introduce our general formalisation for anomaly detection using generative models. We then describe our modified SFDD algorithm and finally discuss the embedding of the framework on the robotic black box introduced in \cite{mitrevski2018}.

    \subsection{Anomaly Detection Using Generative Models: General Formulation}

    We assume that we are given a sequence $\vec{y}$ of $n$ measurements from a given system variable, namely
    \begin{equation}
        \vec{y} = \{ y_1, y_2, ..., y_n \}
    \end{equation}
    We can then define anomaly detection as the problem of finding a set of intervals $F$
    \begin{equation}
        F = \{ [t_i, t_j] \; | \; 1 \leq i, j \leq n, i < j \}
    \end{equation}
    during which the measurements deviate compared to a sequence of nominal observed measurements. Formally, we assume that $\vec{y}$ follows an unknown density $f$ with some additive noise $\epsilon$
    \begin{equation}
        \vec{y} \sim f(\cdot) + \epsilon
    \end{equation}
    where no assumptions have been made on the nature of $f$. We then consider a sample $\vec{y}_{t-k:t}$ of $k$ measurements, where $k$ is a predefined window size, to be nominal if
    \begin{equation*}
        \vec{y}_{t-k:t} \sim f(\cdot) + \epsilon
    \end{equation*}
    Under this formalism, the objective is to learn a model $M$ that represents the unknown data distribution $f$ describing the nominal measurements; $M$ can then be used for verifying whether measurements are likely to follow the distribution. In particular, given $M$ and $\vec{y}_{t-k:t}$, we use a residual generation and comparison paradigm. We define a residual $r$ to be the \emph{dissimilarity between $\vec{y}_{t-k:t}$ and the measurements predicted by $M$}, such that the larger the dissimilarity becomes, the more likely it is that $\vec{y}_{t-k,t}$ is anomalous. Given a dissimilarity measure $d$ and a sample $\vec{m}$ drawn from $M$, we calculate a residual as
    \begin{equation}
        r = d(\vec{y}_{t-k,t}, \vec{m})
        \label{eq:residual_generation}
    \end{equation}
    Using a predefined threshold $\delta$, we can classify $\vec{y}_{t-k,t}$ as
    \begin{equation}
        \left\{
            \begin{array}{ll}
                \text{nominal}, & \text{if } r \leq \delta \\
                \text{faulty} & \text{otherwise}
            \end{array}
        \right.
        \label{eq:residual_comparision}
    \end{equation}

    We use a restricted Boltzmann machine (RBM) to represent $M$ since it fits our main requirements for the model, namely (i) it allows learning $f$ in an unsupervised fashion and, (ii) due to it being a generative model, can be used for sampling from the distribution. Similar to \cite{christensen2008}, temporal information is encoded into the models by using a measurement window as an input to the model. As a distance function, we use the Hellinger distance\footnote{We use the Hellinger distance instead of e.g. the Kullback-Leibler (KL) divergence since the Hellinger distance is symmetric. Conventional distance metrics, such as the L2 norm, were experimentally found to be inappropriate for residual generation.} \cite{nguyen2009}, which measures the discrepancy between two probability distributions and is defined as
    \begin{equation}
        d^2(\vec{a}, \vec{b}) = \frac{1}{2}\sum_{i=1}^{n}\left( \sqrt{a_i} - \sqrt{b_i} \right)^2
    \end{equation}
    for discrete distributions, where $\vec{a}$ and $\vec{b}$ are the measurements that are being compared. It should be noted that if raw sensor measurements are used to represent $\vec{a}$ and $\vec{b}$, $\vec{a}$ and $\vec{b}$ are not valid probability distributions, so we actually abuse the Hellinger distance here.

    \subsection{Dependency Anomaly Detection and Fault Diagnosis}

    Learning the measurement distribution of individual variables as described above would be useful for following the variables' individual trends; however, in line with the SFDD method, we are interested in learning dependency models between variable pairs and monitoring anomalies in those dependencies. As illustrated in Fig. \ref{fig:schema_overview}, we use the following procedure for modelling, monitoring, and subsequently diagnosing faults:
    \begin{enumerate}
        \item We identify the set of correlated pairs of sensors given a set of nominal measurements
        \item For each pair of correlated sensors, a dependency model that encodes the nominal distribution of sliding window correlations is learned
        \item During online operation, we monitor the dependency distribution by comparing samples from the model with the observed correlations
        \item For any anomalous dependencies, conflict sets are generated and used in a subsequent diagnosis step
    \end{enumerate}
    This section describes each of these steps in more detail.

    \subsubsection{Identification of Correlated Sensor Pairs}

    Assuming we are given a structural model of a system, a set $S$ of sensors, and a data set $X$ of nominal measurements, we first identify the set $C$ of correlated sensors as defined in equation \ref{eq:correlated_sensors}. Let $\rho_{S_i, S_j}$ be the correlation between $S_i$ and $S_j$ based on the measurements in $X$. Just as in \cite{khalastchi2018}, we use the Pearson correlation coefficient for identifying correlated sensors; however, the coefficient itself is undefined for constant signals since their variance is zero, which is why we use the following modified definition:
    \begin{equation}
        \rho(\vec{x}, \vec{y}) =
        \left\{
            \begin{array}{ll}
                \frac{\cov(\vec{x}, \vec{y})}{\sigma_i \sigma_j}, & \sigma_i, \sigma_j > 0 \\
                1, & \sigma_i, \sigma_j = 0 \\
                0, & \sigma_i = 0 \; \text{xor} \; \sigma_j = 0
            \end{array}
        \right.
        \label{eq:correlation_coefficient}
    \end{equation}
    where $\vec{x}$ and $\vec{y}$ are sequences of measurements, $\cov$ is the covariance between $\vec{x}$ and $\vec{y}$, and $\sigma$ is the standard deviation. We then have
    \begin{equation}
        \corr(S_i, S_j) =
        \left\{
            \begin{array}{ll}
                1, & \rho_{S_i, S_j} = \rho(\vec{x}_i, \vec{x}_j) > \kappa \\
                0 & \text{otherwise}
            \end{array}
        \right.
    \end{equation}
    for a predefined threshold value $\kappa$, where $\vec{x}_i$ and $\vec{x}_j$ are the measurement sequences of $S_i$ and $S_j$ in $X$ respectively.

    \subsubsection{Learning Dependency Models}

    Given $C$, we learn a generative model $M_{i,j}$ for each pair of correlated sensors $S_i$ and $S_j$. The model learning process is performed in an offline fashion, such that \emph{each $M_{i,j}$ encodes the distribution of the nominal dependency state between $S_i$ and $S_j$}.

    We encode the dependency between two sensors by the correlation between sliding windows extracted from $\vec{x}_i$ and $\vec{x}_j$. Let $k$ be the sliding window size; we then split $\vec{x}_i$ and $\vec{x}_j$ into \emph{overlapping sliding windows} of size $k$\footnote{There are $n-k+1$ such windows in total.} and calculate the correlation between the windows, such that we use the modified correlation coefficient defined in equation \ref{eq:correlation_coefficient}. Calculating the correlation for all windows of size $k$ results in a sequence $\vec{c}_{i,j}$ of windowed correlations between the measurements of $S_i$ and $S_j$.

    For training $M_{i,j}$, we use a sliding window of size $s$\footnote{In other words, we calculate the correlation between pairs of sensors using windows of size $k$, and then use a window of size $s$ of consecutive correlations as an input to each $M_{i,j}$.} with values from $\vec{c}_{i,j}$; \emph{this means that $M_{i,j}$ encodes the distribution of correlations between the measurements of $S_i$ and $S_j$}.

    \subsubsection{Anomaly Detection Using the Dependency Models}

    After learning $M_{i,j}$, we calculate a threshold $\delta_{i,j}$ as
    \begin{equation}
        \delta_{i,j} = \mu_{i,j} + w\sigma_{i,j}
        \label{eq:detection_threshold}
    \end{equation}
    where $\mu_{i,j}$ is the mean residual calculated on the training measurements and $\sigma_{i,j}$ is the standard deviation of the training residuals, and $w \in \mathbb{N}$ is a multiple of the standard deviation. During online operation, we generate a sample $\vec{m_{i,j}}$ given the current input and calculate a residual $r$ as in equation \ref{eq:residual_generation}. The decision about the nominality of the observation then proceeds as in equation \ref{eq:residual_comparision}.

    \subsubsection{Fault Diagnosis}

    For fault diagnosis, just as in \cite{khalastchi2018}, we use the traditional formalisation of DeKleer and Williams \cite{dekleer1987} and Reiter \cite{reiter1987}. We create a conflict set for each of pair of components $S_i$ and $S_j$ for which $r_{i,j}$ exceeds $\delta_{i,j}$; this gives rise to a collection of conflict sets $CS$. Given $CS$, we apply the HS-DAG algorithm \cite{greiner1989} for finding diagnoses using the implementation of the algorithm by Quaritsch and Pill \cite{quaritsch2014}.

    \subsection{Robotic Black Box Application}
    \label{sec:robotic_bb}

    The learning-based framework described here is designed so that it can be used on a robotic black box as described in \cite{mitrevski2018}. The black box continuously logs data during the operation of a robot, where data from different data sources may be logged at different frequencies; however, as described above, the models $M_{i,j}$ require aligned measurements in order for the data correlations to be of any meaningful value. To resolve this issue, the black box needs to log a measurement only if its value changes significantly\footnote{Where the significance level can be set differently for different variables.} compared to its previous value; the measurement is otherwise considered constant. If this condition on the logged data is satisfied, correlations can be calculated even when correlated measurements are observed at different frequencies.

    Another important aspect that needs to be considered is the \emph{operating mode} in which the correlation models are used. Each $M_{i,j}$ is trained with data collected during nominal operation, but depending on the context in which the data are collected, the model might only be usable in certain operating modes (for instance, a robot moving over a smooth floor and moving over an uneven surface). As in \cite{fox2007}, we assume that dedicated models can be created for different operating modes and that the appropriate models will be used depending on the context in which the robot is operating \cite{blanke2006}.

    \section{Experimental Analysis}
    \label{sec:experiments}

\begin{figure}[htp]
        \centering
        \includegraphics[scale=0.25]{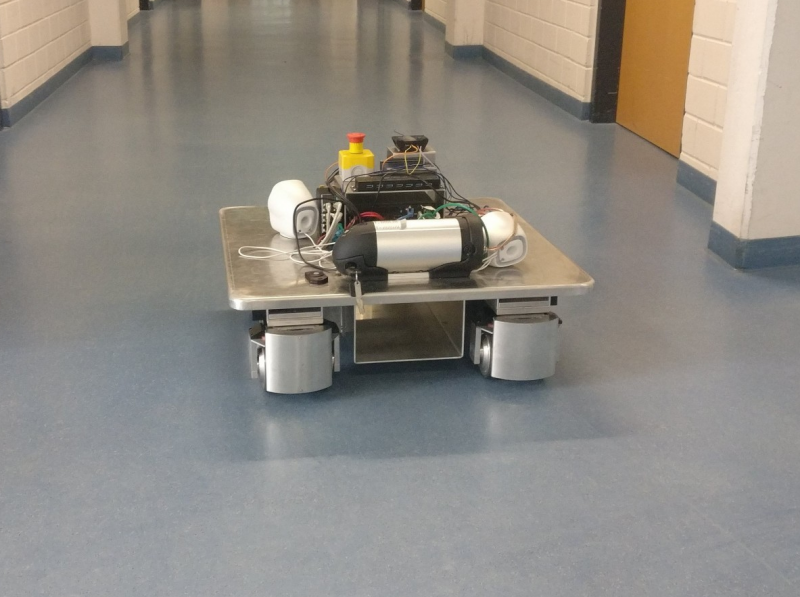}
        \caption{ROPOD platform with four smart wheels}
        \label{fig:ropod}
    \end{figure}

    To show the feasibility of our proposed modification to the SFDD method, we analyse our framework on the ROPOD\footnote{ROPOD is a Horizon 2020 project: \url{http://cordis.europa.eu/project/rcn/206247_en.html}} platform, which is shown in Fig. \ref{fig:ropod}.
    This robot was developed for logistics applications, such that the base is equipped with four so-called \emph{smart wheels}, which are omnidirectional wheels made up of two standard wheels with a caster offset. The wheels themselves provide different sensor measurements\footnote{The wheels receive commands and send feedback through an EtherCAT communication channel.}, particularly current and voltage, wheel velocities, IMU measurements, as well as atmospheric pressure and temperature. The platform itself also uses a 3D lidar for distance measurements; the lidar is also used when the robot needs to attach itself to carts.

\begin{figure*}[tp]
        \begin{subfigure}[t]{0.475\linewidth}
            \centering
            \includegraphics[scale=0.185]{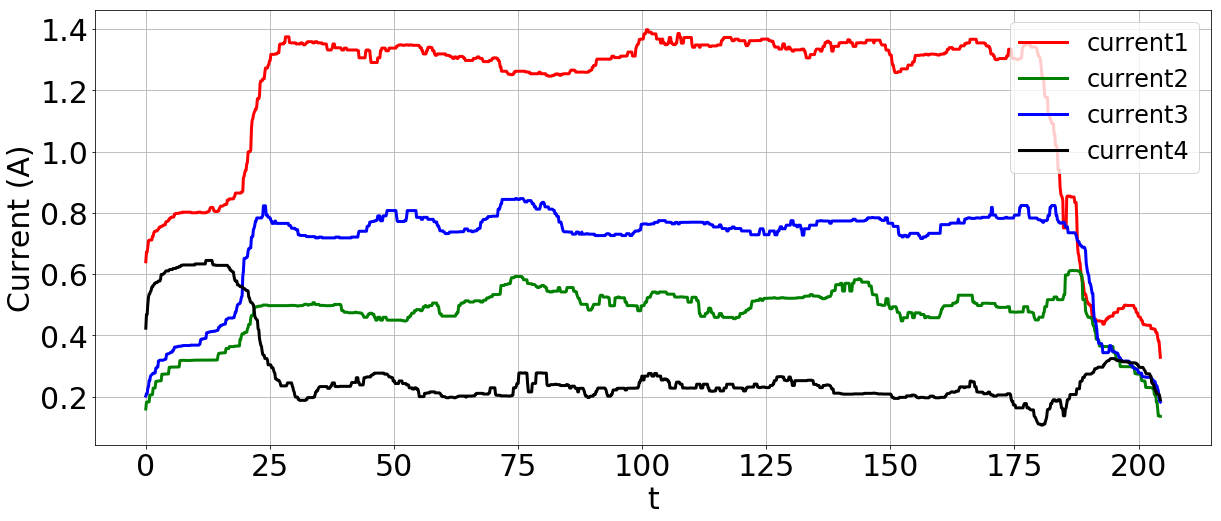}
            \caption{Fault-free current measurements}
            \label{fig:measurements_fault_free}
        \end{subfigure}
        \hspace{0.05\textwidth}
        \begin{subfigure}[t]{0.475\linewidth}
            \centering
            \includegraphics[scale=0.185]{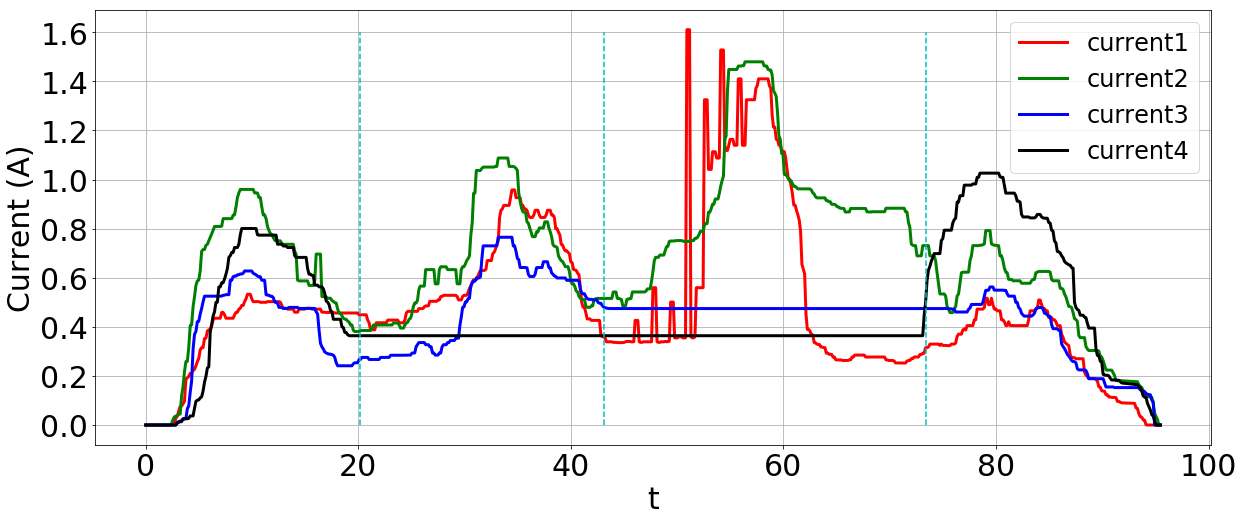}
            \caption{Current measurements with faults in wheels 3 and 4}
            \label{fig:measurements_faulty}
        \end{subfigure}
        \caption{Current measurements from the smart wheels. In the plot of the faulty data, the vertical dotted lines represent events in which faults were first introduced and then removed: the communication of wheel 4 and then 3 was cut off, which are the first two events, while the third event is when the wheels were reconnected back.}
        \label{fig:measurements}
    \end{figure*}

    To simplify the presentation, we constrain our analysis to the current measurements from the smart wheels.\footnote{The code and data for reproducing the results presented in this section can be found at \url{https://github.com/alex-mitrevski/generative-model-fdd}} For training correlation models, we collected data by moving the robot around our university building (as in Fig. \ref{fig:ropod}) with a joypad. The current measurements from this run are shown in Fig. \ref{fig:measurements_fault_free}. For later testing, we collected another data set in which the communication lines of two wheels were cut off one after the other and the wheels were then reconnected back, such that the robot was moved with a joypad both before disconnecting the wheels and while they were disconnected. The current measurements in the faulty data set are shown in Fig. \ref{fig:measurements_faulty}. Since the current measurements are generally noisy, we used a median filter for visualisation, but also as a preprocessing step before learning the models.

\begin{figure*}[tp]
        \begin{subfigure}{0.475\linewidth}
            \centering
            \includegraphics[scale=0.38]{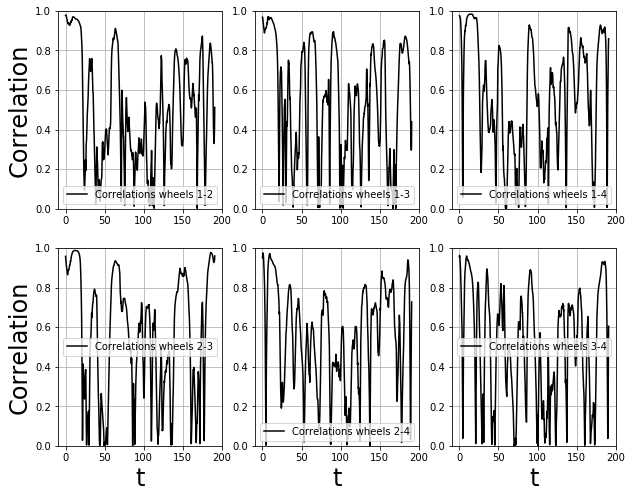}
            \caption{Pairwise correlations on the fault-free data}
            \label{fig:corr_fault_free}
        \end{subfigure}
        \hspace{0.05\textwidth}
        \begin{subfigure}{0.475\linewidth}
            \centering
            \includegraphics[scale=0.38]{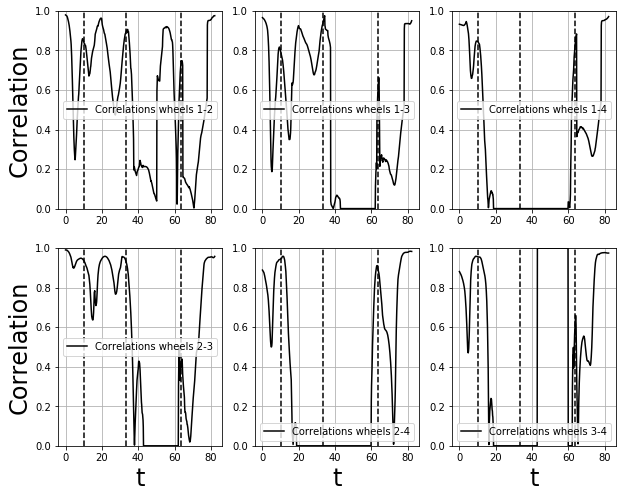}
            \caption{Pairwise correlations on the faulty data}
            \label{fig:corr_faulty}
        \end{subfigure}
        \caption{Pairwise correlations on the fault-free and faulty data. In the plot of the faulty data, the vertical dashed lines represent events that introduce and then remove a fault.}
        \label{fig:corr}
    \end{figure*}

    Given the data set from the nominal operation, we first identified the set of correlated measurements using $\kappa = 0.5$ as a correlation threshold. In this case, all four current measurements are correlated to each other, such that the pairwise correlations of the measurements using a sliding window size $k = 100$ are shown in Fig. \ref{fig:corr_fault_free}. Similarly, the correlations on the faulty data set are visualised in Fig. \ref{fig:corr_faulty}. The models $M_{i, j}$, $1 \leq i < 4$, $ i < j \leq 4$ were trained with the correlations on the fault-free data.\footnote{Since we have binary network units, we normalise the network inputs to lie between $0$ and $1$, using the range of current measurements observed in the nominal data as a normalisation factor.} For each $M_{i,j}$, we used $|V| = 10$ and $|H| = 20$ and CD-$1$ for updating the weights, such that the models were trained for $30$ learning epochs.\footnote{We note that $|V| = s$, the size of the sliding window of values from $\vec{c}_{i,j}$.} We also tried other values of $|V|$ and $|H|$, but since our analysis is not focused on finding the most optimal training parameters, all results below are shown for the previously mentioned values.\footnote{In addition, other values for the sizes of the visible and hidden layer did not significantly affect the results, at least not in this simple use case.}

\begin{figure*}[tp]
        \begin{subfigure}{0.475\linewidth}
            \centering
            \includegraphics[scale=0.38]{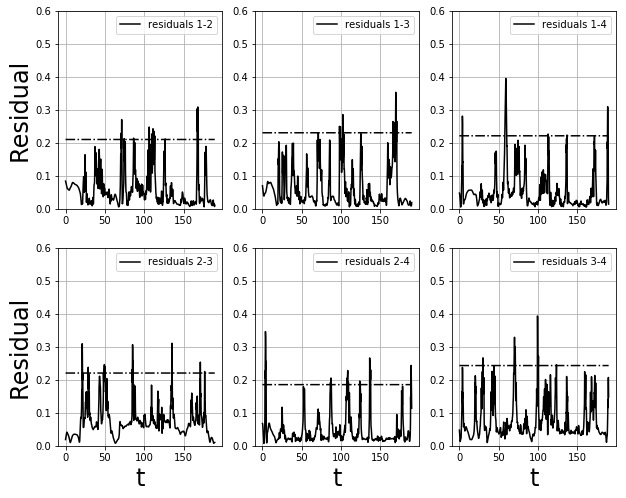}
            \caption{Residuals on the fault free data}
            \label{fig:residuals_fault_free}
        \end{subfigure}
        \hspace{0.05\textwidth}
        \begin{subfigure}{0.475\linewidth}
            \centering
            \includegraphics[scale=0.38]{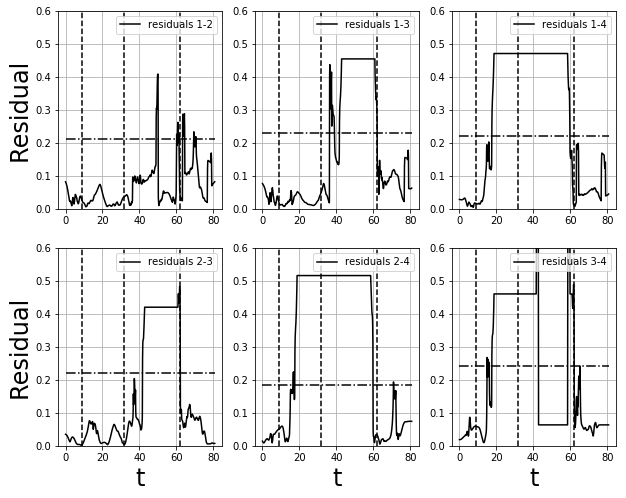}
            \caption{Residuals on the faulty data}
            \label{fig:residuals_faulty}
        \end{subfigure}
        \caption{Residuals on the fault-free and faulty data when using a Restricted Boltzmann Machine. The horizontal dot-dashed lines represent the residual thresholds for each model. In the plot of the faulty data, the vertical dashed lines represent events that introduce and then remove a fault.}
        \label{fig:residuals}
    \end{figure*}

    Once the correlation models were learned, we calculated the thresholds $\delta_{i,j}$ for each model according to equation \ref{eq:detection_threshold} using $w = 3$ as a safe choice. As a sanity check, we show the residuals on the nominal data set in Fig. \ref{fig:residuals_fault_free}. As can be seen there, the models do occasionally raise false alarms since the residuals cross the detection thresholds, but these are transient detections that could be avoided by smoothing the detections, for instance using a hidden Markov model \cite{fox2007}. The more interesting residuals are the ones on the faulty data set, which are depicted in Fig. \ref{fig:residuals_faulty}. As can be seen there, the models were able to reliably detect the injected faults; the residuals then went down within the normal range when the faults were removed. One observation that can be made here is that the detections are slightly delayed, which is however to be expected since the correlations are calculated over a sliding window whose size affects the observed delay (as mentioned before, we used a conservative value of $k = 100$ in our experiments).

    To evaluate the detection results quantitatively, we use the common precision and recall metrics:
    \begin{align*}
        \operatorname{precision} &= \frac{TP}{TP + FP} \\
        \operatorname{recall} &= \frac{TP}{TP + FN}
    \end{align*}
    where $TP$ is the number of true positive detections over all time steps, $FP$ is the number of false alarms, and $FN$ is the number of missed detections, such that it should be noted that these were calculated jointly over all pairwise models. The results shown in Fig. \ref{fig:residuals} evaluate to $88.6\%$ precision and $75.6\%$ recall. The false alarms affecting the precision rate are particularly visible for the correlation model between the currents of wheels 1 and 2; on the other hand, the delayed detections caused by the sliding window size result in a relatively low recall rate, but this is expected given the conservative value of $k$.

    As mentioned in the previous section, the residuals calculated from the models produce conflict sets consisting of pairs of components. In the case of both wheel 3 and wheel 4 being disconnected, the collection of conflict sets is given as follows:
    \begin{equation*}
        X = \{ \{ c_1, c_3 \}, \{ c_1, c_4 \}, \{ c_2, c_3 \}, \{ c_2, c_4 \}, \{ c_3, c_4 \} \}
    \end{equation*}
    If we constrain ourselves to diagnoses of maximum cardinality $2$, applying HS-DAG on the above collection of conflict sets results in the only diagnosis $\{ c_3, c_4 \}$, which is clearly also the correct diagnosis in this case. If we allow cardinality $3$ for the diagnoses, we also obtain $\{ c_1, c_2, c_3 \}$ and $\{ c_1, c_2, c_4 \}$ as potential diagnoses.

\begin{figure*}[tp]
        \begin{subfigure}{0.475\linewidth}
            \centering
            \includegraphics[scale=0.38]{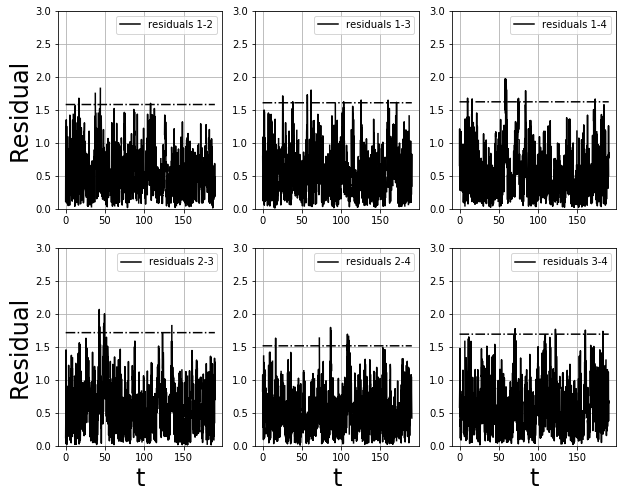}
            \caption{Residuals on the fault free data}
            \label{fig:residuals_fault_free_gmm}
        \end{subfigure}
        \hspace{0.05\textwidth}
        \begin{subfigure}{0.475\linewidth}
            \centering
            \includegraphics[scale=0.38]{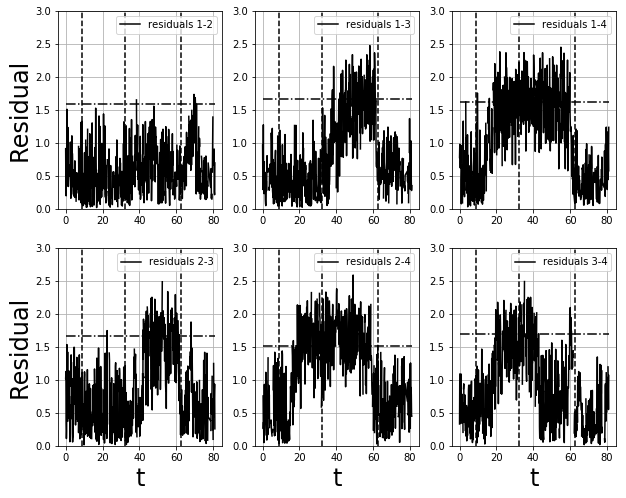}
            \caption{Residuals on the faulty data}
            \label{fig:residuals_faulty_gmm}
        \end{subfigure}
        \caption{Residuals on the fault-free and faulty data when using a Gaussian Mixture Model. The horizontal dot-dashed lines represent the residual thresholds for each model. In the plot of the faulty data, the vertical dashed lines represent events that introduce and then remove a fault.}
        \label{fig:residuals_gmm}
    \end{figure*}

    For the purposes of completeness, we also used an alternative representation of the correlation models $M_{i,j}$, namely we represented them by Gaussian Mixture Models (GMMs) instead of RBMs. The results of applying GMMs for residual generation on the fault-free and faulty data sets are depicted in Fig. \ref{fig:residuals_gmm}. We used five mixture components in this case; increasing the number of components did not seem to affect the results significantly. As can be seen here, GMMs are also able to identify the trend in the data - we obtain $92.1\%$ precision when using a GMMs as a generative model, which is slightly better than the precision of the RBM - but the residuals are generally more noisy than in the case of the RBMs and thus require post-processing in order to be of practical value. This is confirmed by the recall on the faulty data obtained when using GMMs, which is only $34.6\%$.

    \section{Discussion and Conclusions}
    \label{sec:discussion}

    This paper discussed a modification of the sensor-based fault detection and diagnosis (SFDD) algorithm by Khalastchi and Kalech \cite{khalastchi2018} based on which the manually specified data modes required by the algorithm are replaced by models of pairwise sliding window correlations between sets of correlated measurements. The distribution of each correlation pair is represented by a generative model, which is used for online residual generation and subsequent conflict generation and diagnosis.

    As is generally the case when a learning-based model is used, we assume that the data used for training the models are representative enough of the nominal operation of a robot. Since nominal data are generally more abundant than faulty data, we see this as an advantage of our method over methods based on discriminative models (e.g. \cite{matsuno2013}) since those also require faulty data for learning; however, if the detection of specific faults is desired, a discriminative model may be more appropriate than a fault-independent generative model.

    There are various aspects that need to be addressed in future work. First of all, as discussed in section \ref{sec:robotic_bb}, different operating modes may require different sets of correlation models since the correlations may change between operating modes; this however necessitates a suitable context transition and potentially recognition model, such as for instance \cite{karg2012}. Related to that, our diagnosis module as discussed here is quite rudimentary and is limited to diagnosing component faults, but could also be extended for diagnosing higher-level execution failures \cite{banerjee2019}. Furthermore, even though our method is not robot-specific, the portability to different robots, for instance to mobile manipulators, needs to be investigated in a follow-up study.\footnote{In our lab, we are planning to use our method on a Toyota HSR mobile manipulator \cite{yamamoto2019} and a KINOVA KORTEX Gen3 arm.} Finally, on a more conceptual note, while the correlation between pairs of sensor measurements does seem to be a suitable feature for anomaly detection, it should be possible to obtain similar results with other features, such as moving averages or sliding window finite differences. This also applies to the distance metric used for residual generation; the Hellinger distance was experimentally found to be suitable for generating residuals, but it should be possible to replicate the results with other distance metrics. Both the features and the distance metric are thus design parameters, which may also be learnable in an end-to-end fashion.

    \section*{Acknowledgments}

    ROPOD is an Innovation Action funded by the European Commission under grant no. 731848 within the Horizon 2020 framework program. We are additionally grateful for the continuous support by the b-it International Center for Information Technology. We would also like to thank Ahmed Abdelrahman for his comments on our manuscript.

\nocite{*}
\bibliographystyle{unsrt}
\bibliography{references}

\end{document}